\documentclass[10pt,twocolumn,letterpaper]{article}

\usepackage{iccv}
\usepackage{times}
\usepackage{epsfig}
\usepackage{graphicx}
\usepackage{amsmath}
\usepackage{amssymb}

\usepackage{bbm}
\usepackage{url}
\usepackage{multirow,makecell,rotating}
\usepackage{booktabs}
\usepackage{caption}
\usepackage{graphbox}
\usepackage{comment}
\usepackage{epsfig}
\usepackage{bm}
\usepackage{soul}
\usepackage{color}
\usepackage{xcolor}
\usepackage{appendix}

\definecolor{citecolor}{HTML}{0071BC}
\definecolor{linkcolor}{HTML}{ED1C24}
\definecolor{highlightcolor}{HTML}{ABCDEF}
\usepackage[pagebackref, breaklinks, colorlinks, letterpaper=true, citecolor=citecolor, linkcolor=linkcolor, bookmarks=false]{hyperref}

\iccvfinalcopy % *** Uncomment this line for the final submission

\newcommand{\model}{Tune-A-Video}

\newcommand{\setting}{One-Shot Video Tuning}

\begin{document}

%%%%%%%%% TITLE
\title{Tune-A-Video: One-Shot Tuning of Image Diffusion Models\\ for Text-to-Video Generation}

\author{Jay Zhangjie Wu$^{1}$\quad Yixiao Ge$^{2}$\quad Xintao Wang$^{2}$\quad Stan Weixian Lei$^{1}$\quad Yuchao Gu$^{1}$ 
\\ 
Yufei Shi$^{1}$\quad Wynne Hsu$^{4}$\quad Ying Shan$^{2}$\quad Xiaohu Qie$^{3}$\quad Mike Zheng Shou$^1$\thanks{}
\\ \vspace{-0.6em} \\ 
$^1$Show Lab, National University of Singapore\quad $^2$ARC Lab,$^3$Tencent PCG \\ 
$^4$School of Computing, National University of Singapore
\\ \vspace{-0.6em} \\ 
\url{https://tuneavideo.github.io}
}

\twocolumn[{
\maketitle
\vspace{-2.4em}
\renewcommand\twocolumn[1][]{#1}
\begin{center}
    \centering
    \includegraphics[width=0.93\textwidth]{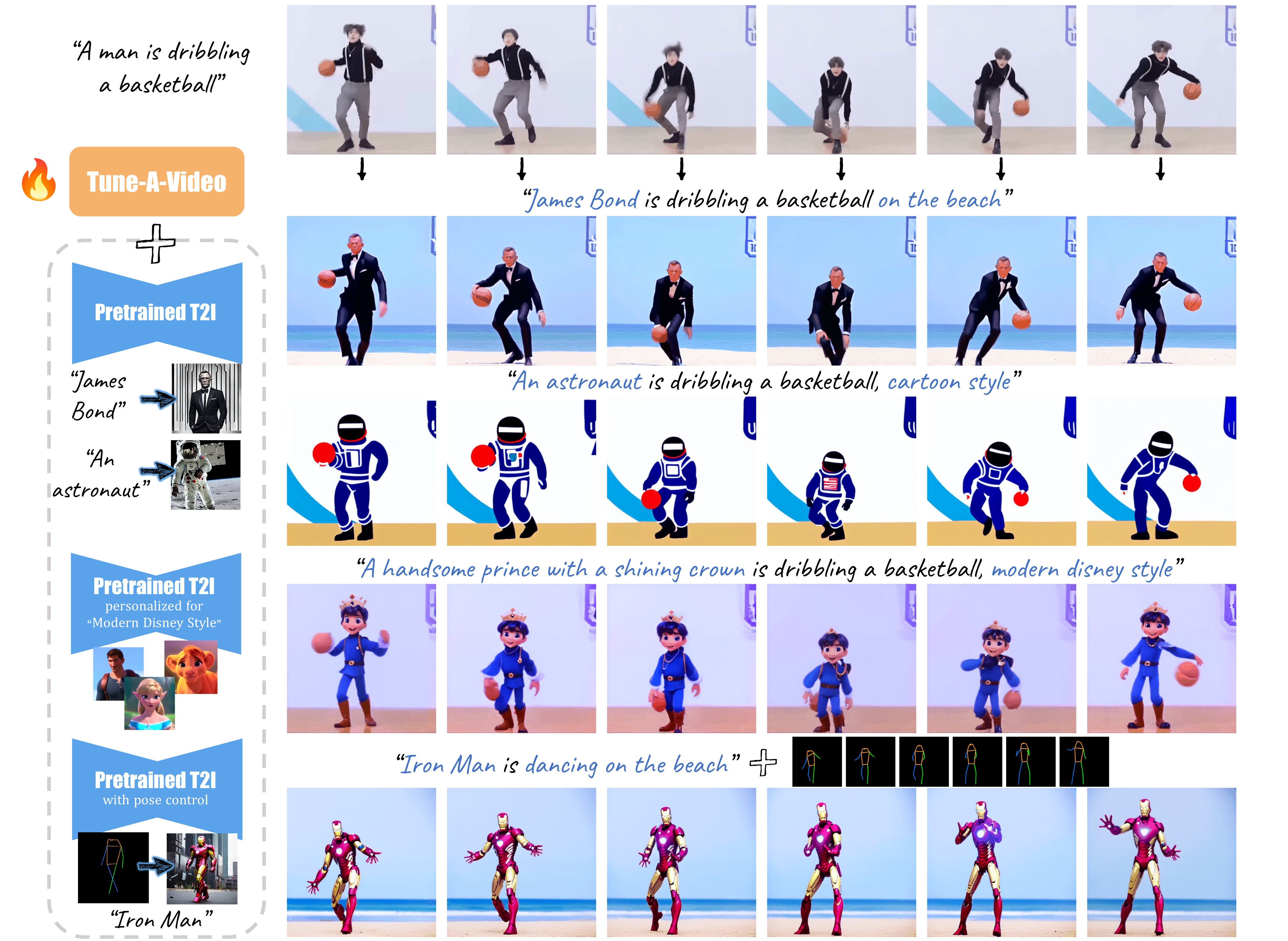}
    % \vspace{-0.2cm}
    \captionof{figure}{
    \textit{\textbf{Tune-A-Video:}} A new method for T2V generation using one text-video pair and pretrained T2I models. 
    }
    \label{fig:teaser}
    % \vspace{-0.8em}
\end{center}
}]

{
  \renewcommand{\thefootnote}%
    {\fnsymbol{footnote}}
  \footnotetext[1]{Corresponding Author.}
}


%%%%%%%%% ABSTRACT
\begin{abstract}
To replicate the success of text-to-image (T2I) generation, recent works employ large-scale video datasets to train a text-to-video (T2V) generator. Despite their promising results, such paradigm is computationally expensive. In this work, we propose a new T2V generation setting---\setting{}, where only one text-video pair is presented. Our model is built on state-of-the-art T2I diffusion models pre-trained on massive image data. We make two key observations: 1) T2I models can generate still images that represent verb terms; 2) extending T2I models to generate multiple images concurrently exhibits surprisingly good content consistency. To further learn continuous motion, we introduce \model{}, which involves a tailored spatio-temporal attention mechanism and an efficient one-shot tuning strategy. At inference, we employ DDIM inversion to provide structure guidance for sampling. Extensive qualitative and numerical experiments demonstrate the remarkable ability of our method across various applications. 
\end{abstract}

%%%%%%%%% BODY TEXT
\section{Introduction}

The large-scale multimodal dataset~\cite{schuhmann2022laion}, consisting of billions of text-image pairs crawled from the Internet, has enabled a breakthrough in Text-to-Image (T2I) generation~\cite{nichol2021glide, ramesh2022hierarchical, ding2022cogview2, singer2022make, saharia2022photorealistic}. To replicate this success in Text-to-Video (T2V) generation, recent works~\cite{singer2022make, ho2022imagen, ho2022video, zhou2022magicvideo, villegas2022phenaki} have extended spatial-only T2I generation models to the spatio-temporal domain. These models generally adopt the standard paradigm of training on large-scale text-video datasets (\eg, WebVid-10M~\cite{bain2021frozen}). Although this paradigm produces promising results for T2V generation, it requires extensive training on large hardware accelerators, which is expensive and time-consuming.

Humans possess the ability to create new concepts, ideas, or things by utilizing their existing knowledge and the information provided to them. For example, when presented a video with a textual description of ``a man skiing on snow", we can imagine how a panda would ski on snow, drawing upon our knowledge of what a panda looks like. As T2I models pretrained with large-scale image-text data already capture knowledge of open-domain concepts, a intuitive question arises: \textit{can they infer other novel videos from a single video example, like humans?} 
A new T2V generation setting is therefore introduced, namely, \setting{}, where only a single text-video pair is used to train a T2V generator. The generator is expected to capture essential motion information from the input video and synthesize novel videos with edited prompts.

\begin{figure}[t!]
    \centering
    \includegraphics[width=0.99\linewidth]{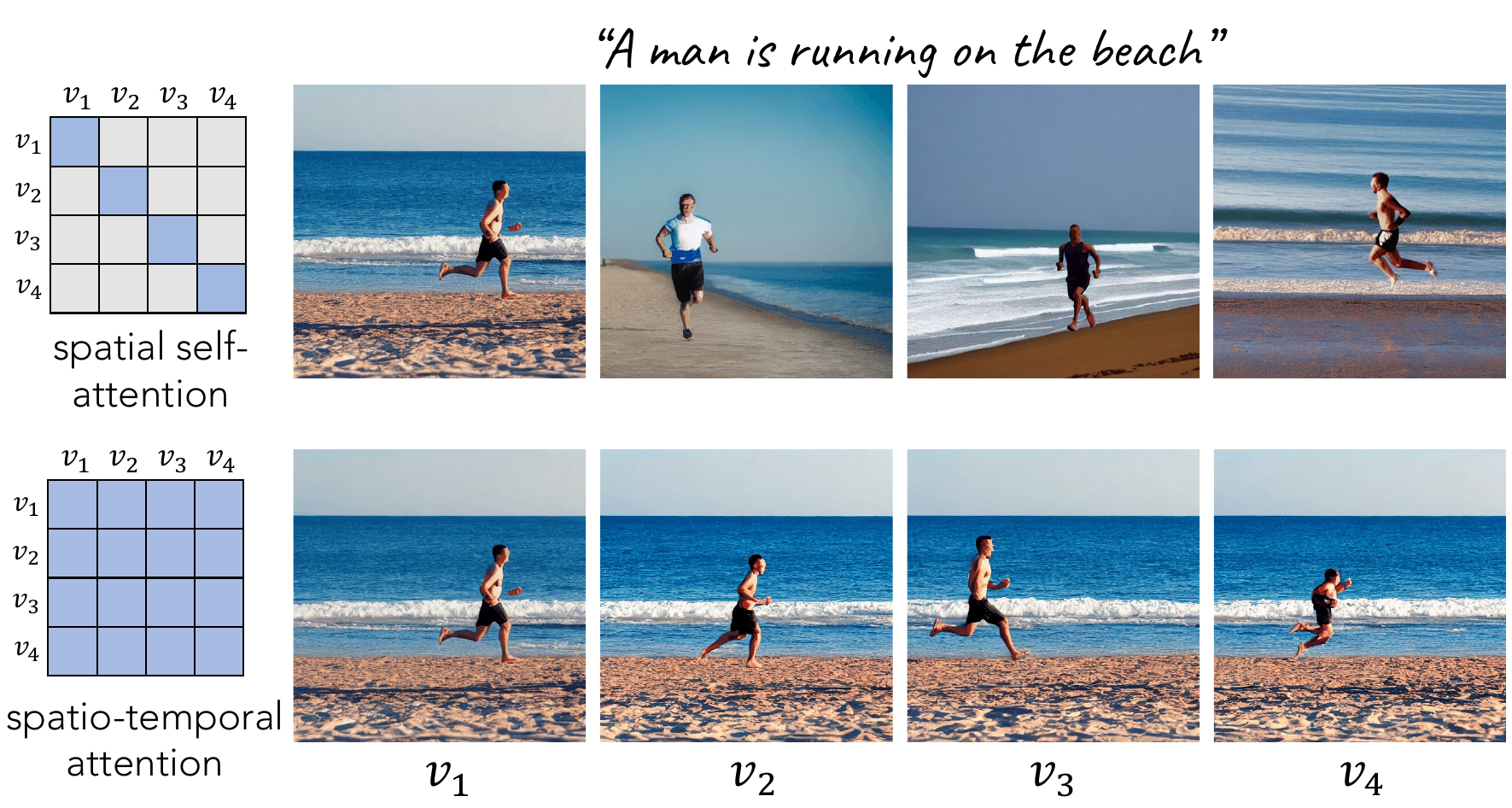}
    \caption{\textit{\textbf{Observations on pretrained T2I models:}} 1) They can generate still images that accurately represent the verb terms. 2) Extending spatial self-attention to spatio-temporal attention produces consistent content across frames.}
    \label{fig:obs}
    \vspace{-1em}
\end{figure}

Intuitively, the key to successful video generation lies in preserving the continuous motion of consistent objects.
So we make the following observations on state-of-the-art T2I diffusion models~\cite{rombach2022high} that motivate our method accordingly.
(1) \textbf{Regarding motion:} T2I models are able to generate images that align well with the text, including the verb terms. For example, given the text prompt ``a man is running on the beach", the T2I models produce the snapshot where a man is running (not walking or jumping), albeit not necessarily in a continuous manner (the first row of Fig.~\ref{fig:obs}). This serves as evidence that T2I models can properly attend to verbs via cross-modal attention for static motion generation. 
(2) \textbf{Regarding consistent objects:} Simply extending the spatial self-attention in the T2I model from one image to multiple images produces consistent content across frames. Taking the same example, when we generate consecutive frames in parallel with extended spatio-temporal attention, the same man and the same beach can be observed in the resultant sequence though the motion is still not continuous (the second row of Fig.~\ref{fig:obs}). This implies that the self-attention layers in T2I models are only driven by spatial similarities rather than pixel positions.

We implement our findings into a simple yet effective method called \model{}.
Our method is based on a simple inflation of state-of-the-art T2I models over spatio-temporal dimension. However, using full attention in space-time inevitably leads to quadratic growth in computation. It is thus infeasible for generating videos with increasing frames. Additionally, employing a naive fine-tuning strategy that updates all the parameters can jeopardize the pre-existing knowledge of T2I models and hinder the generation of videos with new concepts. To tackle these problems, we introduce a sparse spatio-temporal attention mechanism that only visits the \textit{first} and the \textit{former} video frame, as well as an efficient tuning strategy that only updates the projection matrices in attention blocks. Empirically, these designs maintain consistent objects across all frames but lack continuous motion. Therefore, at inference, we further seek structure guidance from input video through DDIM inversion, which is a reverse process of DDIM sampling~\cite{song2020denoising}. With the inverted latent as initial noise, we produce temporally-coherent videos featuring smooth movement. Notably, our method is inherently compatible with exiting personalized and conditional pretrained T2I models, such as DreamBooth~\cite{ruiz2022dreambooth} and T2I-Adapter~\cite{mou2023t2i}, providing a personalized and controllable user interface.

We showcase remarkable results of \model{} across a wide range of applications for text-driven video generation (see Fig.~\ref{fig:teaser}). We compare our method against the state-of-the-art baselines through extensive qualitative and quantitative experiments, demonstrating its superiority. 
In summary, our key contributions are as follows: 
\begin{itemize}
    \itemsep0em 
    \item We introduce a new setting of \setting{} for T2V generation, which eliminates the burden of training with large-scale video datasets.
    \item We present \model{}, which is the first framework for T2V generation using pretrained T2I models.
    \item We propose efficient attention tuning and structural inversion that significant improve temporal consistency.
    \item We demonstrate remarkable results of our method through extensive experiments. 
\end{itemize}

\section{Related Work}

Our work lies in the intersection of several fields: diffusion models and methods for image/video generation from text prompts, text-driven editing of a real image/video, and generative models trained on a single video. 
Here we provide a brief overview of the key accomplishments in each field, highlighting their connections and differences from our proposed method.

\vspace{-1em}
\paragraph{Text-to-Image diffusion models.}

% Transformer-based models have shown remarkable performance in text-to-image generation. DALL-E~\cite{ramesh2021zero} uses sequence-to-sequence translation to generate images from textual descriptions by representing text tokens as discrete image embeddings learned by a discrete VQ-VAE~\cite{van2017neural}. Parti~\cite{yu2022scaling} uses an encoder-decoder architecture with an advanced image tokenizer, ViT-VQGAN~\cite{yu2021vector}, to generate images. CogView2~\cite{ding2022cogview2} achieves competitive results using hierarchical transformers and local parallel auto-regressive generation, while Make-A-Scene~\cite{gafni2022make} employs domain-specific knowledge to improve controllability.
% Recently, Denoising Diffusion Probabilistic Models (DDPMs)~\cite{ho2020denoising} have gained popularity in text-to-image generation.

Text-to-Image (T2I) generation has been studied extensively, in past years many of the models were based on transformers~\cite{ramesh2021zero, yu2022scaling, yu2021vector, ding2022cogview2, gafni2022make}. Several T2I generative models~\cite{nichol2021glide, saharia2022photorealistic, gu2022vector, rombach2022high} have recently adopted diffusion models~\cite{ho2020denoising}. GLIDE~\cite{nichol2021glide} proposes classifier-free guidance~\cite{ho2022classifier} in the diffusion model to improve image quality, while DALLE-2~\cite{ramesh2022hierarchical} improves text-image alignments using CLIP~\cite{radford2021learning} feature space. Imagen~\cite{saharia2022photorealistic} uses cascaded diffusion models for high definition video generation, and subsequent works like VQ-diffusion~\cite{gu2022vector} and Latent Diffusion Models (LDMs)~\cite{rombach2022high} operate in the latent space of an autoencoder to improve training efficiency. Our method builds on LDMs, by inflating the 2D model to spatio-temporal domain in latent space.

\vspace{-1em}
\paragraph{Text-to-Video generative models.}

While there have been significant advancements in T2I generation, generating videos from text is still lagging behind due to the scarcity of high-quality, large-scale text-video datasets, and the inherent complexity of modeling temporal consistency and coherence. 
Early works~\cite{mittal2017sync, pan2017create, marwah2017attentive, li2018video, gupta2018imagine, liu2019cross} primarily focus on generating videos in simple domains, such as moving digits or specific human actions.  Recently, GODIVA~\cite{van2017neural} is the first model to utilize 2D VQ-VAE and sparse attention for T2V generation, which allows for more realistic scenes. NÜWA~\cite{wu2022nuwa} expands upon GODIVA by presenting a unified representation for various generation tasks through a multitask learning approach. To further enhance T2V generation performance, CogVideo~\cite{hong2022cogvideo} is developed by incorporating additional temporal attention modules on top of a pre-trained T2I model, CogView2~\cite{ding2022cogview2}.

To replicate the success of T2I diffusion models, Video Diffusion Models (VDM)~\cite{ho2022video} uses a space-time factorized U-Net with joint image and video data training. Imagen Video~\cite{ho2022imagen} improves VDM using cascaded diffusion models and v-prediction parameterization to generate high definition videos. Make-A-Video~\cite{singer2022make} and MagicVideo~\cite{zhou2022magicvideo} share similar motivations and aim to transfer progress from T2I generation to T2V generation. 
Although current T2V generative models have shown impressive results, their success heavily rely on being trained using extensive video data. In contrast, we present a new framework for T2V generation via an efficient tuning of pre-trained T2I diffusion models on one text-video pair.

\vspace{-1em}
\paragraph{Text-driven video editing.}

Recent diffusion-based image editing models~\cite{meng2021sdedit, hertz2022prompt, couairon2022diffedit, wu2022unifying, kawar2022imagic, tumanyan2022plug} can process each individual frame in a video, but this produces inconsistency between frames due to the lack of temporal awareness in the model. Text2Live~\cite{bar2022text2live} allows some texture-based video editing using text prompts, but struggles to accurately reflect the intended edits due to its dependence on Layered Neural Atlases~\cite{kasten2021layered}. Moreover, generating a neural atlas typically takes about 10 hours, whereas our approach only requires a 10-minute training per video and can sample a video in just 1 minute.
Two concurrent works, Dreamix~\cite{molad2023dreamix} and Gen-1~\cite{esser2023structure}, both utilize the video diffusion model (VDM) for video editing purposes. Although their impressive outcomes, it is worth noting that the VDMs are computationally demanding and necessitate large-scale captioned images and videos for training. Additionally, their training data and pre-trained models are not publicly accessible.

\vspace{-1em}
\paragraph{Generation from a single video.}

Single-video GANs~\cite{arora2021singan-gif,gur2020hierarchical} generate new videos of similar appearance and dynamics to the input video. However, these GAN-based methods are limited in computation time (e.g., HPVAE-GAN~\cite{gur2020hierarchical} takes 8 days to train on a short video of 13 frames), and thus are impractical and unscalable to some extent. Patch nearest-neighbour methods~\cite{haim2022diverse} perform video generation of higher quality while reducing computation expense by orders of magnitude. However, they are limited in generalization, and therefore can only handle tasks where it is natural to ``copy" parts of the input video. Lately, SinFusion~\cite{nikankin2022sinfusion} adapts diffusion models to single-video tasks, and enables autoregressive video generation with improved motion generalization capabilities; however, it is still incapable of producing videos that contains novel semantic contexts. 

\begin{figure}[t!]
    \centering
    \includegraphics[width=0.99\linewidth]{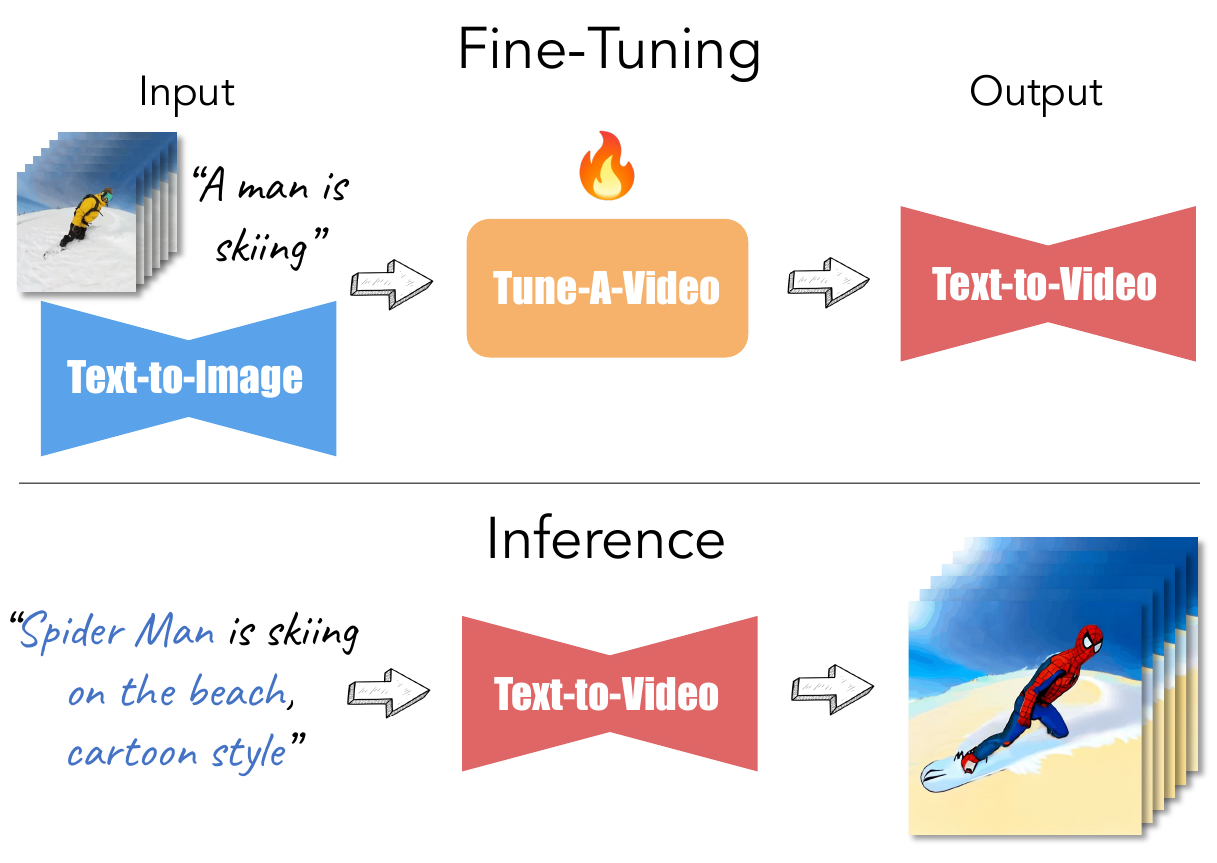}
    \caption{\textit{\textbf{High-level overview of \model{}.}} Given a captioned video, we finetune a pre-trained T2I model (\eg, Stable Diffusion) for T2V modeling. During inference, we generate novel videos that represent the edits in text prompt while preserving the temporal consistency of input video.}
    \label{fig:overview}
    \vspace{-1em}
\end{figure}

\begin{figure*}[t!]
    \centering
    \includegraphics[width=0.99\linewidth]{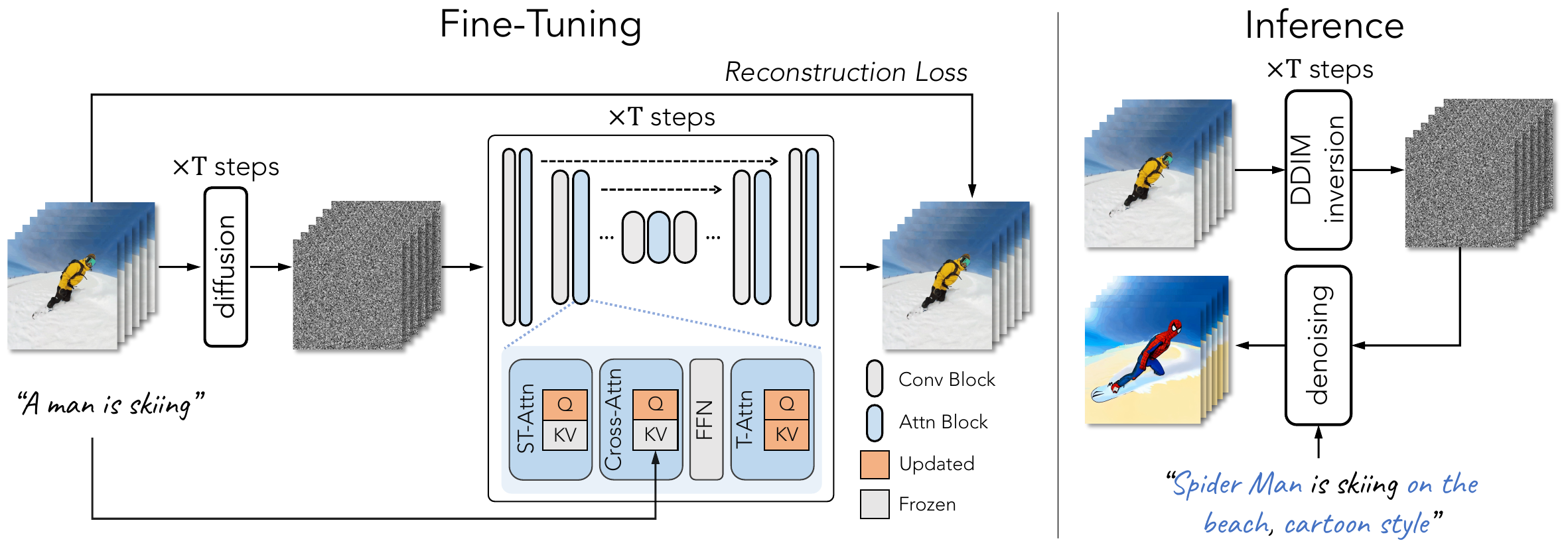}
    \caption{\textit{\textbf{Pipeline of \model{}:}} 
    Given a text-video pair (\eg, ``a man is skiing") as input, our method leverages the pretrained T2I diffusion models for T2V generation. During fine-tuning, we update the projection matrices in attention blocks using the standard diffusion training loss. During inference, we sample a novel video from the latent noise inverted from the input video, guided by an edited prompt (\eg, ``Spider Man is surfing on the beach, cartoon style").}
    \label{fig:arch}
    \vspace{-1em}
\end{figure*}

\begin{figure}[t!]
    \centering
    \includegraphics[width=0.99\linewidth]{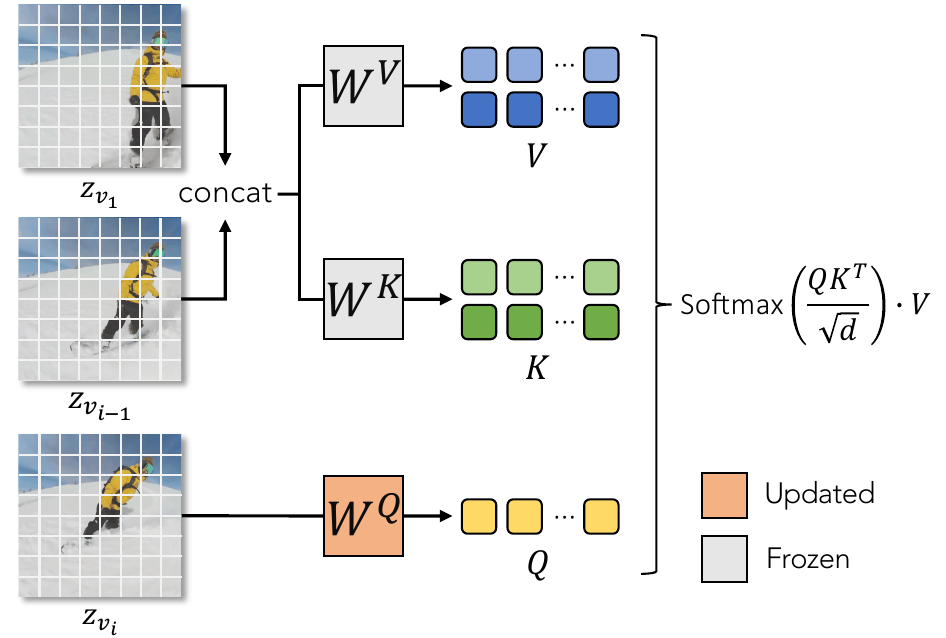}
    \caption{\textit{\textbf{Illustration of our ST-Attn:}} Latent features of frame $v_i$, previous frames $v_{i-1}$ and $v_1$ are projected to query $Q$, key $K$ and value $V$. Output is a weighted sum of the values, weighted by the similarity between the query and key features. We highlight the updated parameter $W^Q$.}
    \label{fig:our-attn}
    \vspace{-1em}
\end{figure}

\begin{figure*}[t!]
    \centering
    \includegraphics[width=0.99\linewidth]{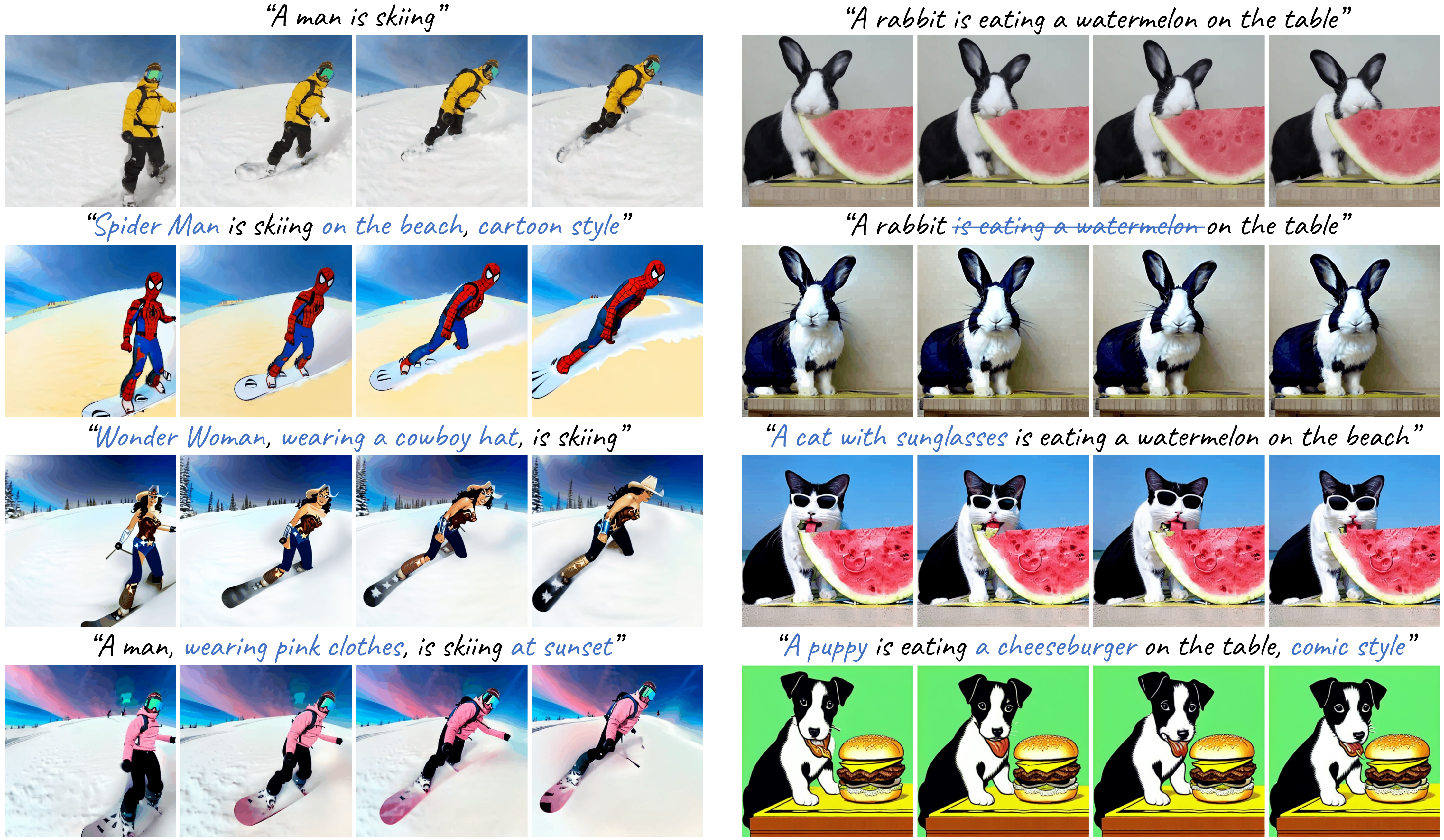}
    \caption{\textit{\textbf{Sample results of our method.}}}
    \label{fig:results}
    \vspace{-1em}
\end{figure*}

\section{Method}

Let $\mathcal{V}=\left\{v_i|i\in[1,m]\right\}$ be a video containing $m$ frames, $\mathcal{P}$ be the source prompt describing $\mathcal{V}$. Our goal is to generate a novel video $\mathcal{V^\ast}$ driven by an edited text prompt $\mathcal{P^\ast}$. For example, consider a video and a source prompt ``a man is skiing", and assume that the user wants to alter the color of the clothes, incorporate a cowboy hat to the skier, or even replace the skier with Spider Man while preserving the motion of the original video. The user can directly modify the source prompt by further describing the appearance of the skier or replacing it with another word. 

An intuitive solution is to train a T2V model on large-scale video datasets, but it is computationally expensive~\cite{singer2022make, ho2022imagen, esser2023structure}. In this paper, we propose a new setting called \setting{} that achieves the same goal using a publicly available T2I model and a single text-video pair. 

Next, we provide a short background of diffusion models in Sec.~\ref{sec:preliminary}, followed by a detailed description of our method in Sec.~\ref{sec:network} and Sec.~\ref{sec:finetune}. An overview of our approach is depicted in Fig.~\ref{fig:overview}.

\subsection{Preliminaries}\label{sec:preliminary}

\paragraph{Denoising diffusion probabilistic models (DDPMs).} DDPMs~\cite{ho2020denoising} are latent generative models trained to recreate a fixed forward Markov chain $x_1, \ldots, x_T$.
Given the data distribution $x_0 \sim q(x_0)$, the Markov transition $q(x_t|x_{t-1})$ is defined as a Gaussian distribution with a variance schedule $\beta_t \in (0, 1)$, that is,
{\small
\begin{equation*}
    q(x_t|x_{t-1}) = \mathcal{N}(x_t; \sqrt{1-\beta_t} x_{t-1}, \beta_t \mathbb{I}), \quad t = 1, \ldots, T.
\end{equation*}}%
By the Bayes' rules and Markov property, one can explicitly express the conditional probabilities $q(x_t|x_0)$ and $q(x_{t-1}|x_t, x_0)$ as
{\small
\begin{align*}
    q(x_t|x_0) & = \mathcal{N}(x_t; \sqrt{\Bar{\alpha}_t} x_0, (1 - \Bar{\alpha}_t) \mathbb{I}), \quad t = 1, \ldots, T, \\
    q(x_{t-1}|x_t, x_0) & = \mathcal{N}(x_{t-1}; \Tilde{\mu}_t(x_t, x_0), \Tilde{\beta}_t \mathbb{I}), \quad t = 1, \ldots, T, \\
    w.r.t.~~~~ &\alpha_t = 1 - \beta_t, ~ \Bar{\alpha}_t = \prod_{s=1}^t \alpha_s, ~ \Tilde{\beta}_t = \frac{1 - \Bar{\alpha}_{t-1}}{1 - \Bar{\alpha}_t} \beta_t, \\
    &\Tilde{\mu}_t(x_t, x_0) = \frac{\sqrt{\Bar{\alpha}_t}\beta_t}{1 - \Bar{\alpha}_t} x_0 + \frac{\sqrt{{\alpha}_t} (1 - \Bar{\alpha}_{t-1})}{1 - \Bar{\alpha}_t} x_t.
\end{align*}
}%
To generate the Markov chain $x_1, \ldots, x_T$, DDPMs leverage the reverse process with a prior distribution $p(x_T) = \mathcal{N}(x_T; 0, \mathbb{I})$ and Gaussian transitions
{\small
\begin{equation*}
    p_\theta(x_{t-1}|x_t) = \mathcal{N}(x_{t-1}; \mu_\theta (x_t, t), \Sigma_\theta(x_t, t)), \quad t = T, \ldots, 1.
\end{equation*}}%
Learnable parameters $\theta$ are trained to guarantee that the generated reverse process is close to the forward process.

To this end, DDPMs follow the variational inference principle by maximizing the variational lower bound of the negative log-likelihood, which has a closed-form given the KL divergence among Guassian distributions. Empirically, these models can be interpreted as a sequence of weight-sharing denoising autoencoders $\epsilon_\theta(x_t, t)$, which are trained to predict a denoised variant of their input $x_t$. The objective can be simplified as $\mathbb{E}_{x, \epsilon \sim \mathcal{N}(0, 1), t} \left[ \| \epsilon - \epsilon_\theta(x_t, t) \|^2_2 \right]. $

\vspace{-1em}
\paragraph{Latent diffusion models (LDMs).} LDMs~\cite{rombach2022high} are newly introduced variants of DDPMs that operate in the latent space of an autoencoder. LDMs consist of two key components. First, an autoencoder~\cite{esser2021taming, van2017neural} is trained with patch-wise losses on a large collection of images, where an encoder $\mathcal{E}$ learns to compress images $x$ into latent representations $z=\mathcal{E}(x)$, and a decoder $\mathcal{D}$ learns to reconstruct the latent back to pixel space, such that $\mathcal{D}(\mathcal{E}(x)) \approx x$. The second component is a DDPM that is trained to remove the noise added to the sampled data. 
For a text-guided LDM, the objective is given by: $\mathbb{E}_{z, \epsilon \sim \mathcal{N}(0, 1), t, c} \left[ \| \epsilon - \epsilon_\theta(z_t, t, c) \|^2_2 \right].$, where $c = \psi(\mathcal{P^\ast})$ is the embedding of textual condition $\mathcal{P^\ast}$.

\subsection{Network Inflation}\label{sec:network}

A T2I diffusion model (\eg, LDM~\cite{rombach2022high}) typically employs a U-Net~\cite{ronneberger2015u}, which is a neural network architecture based on a spatial downsampling pass followed by an upsampling pass with skip connections. It is composed of stacked 2D convolutional residual blocks and transformer blocks. Each transformer block consists of a spatial self-attention layer, a cross-attention layer, and a feed-forward network (FFN). The spatial self-attention leverages pixel locations in feature maps for similar correlation, while the cross-attention considers correspondence between pixels and conditional inputs (\eg, text).
Formally, given latent representation $z_{v_i}$ of video frame ${v_i}$, the spatial self-attention mechanism~\cite{vaswani2017attention} implements $\mathrm{Attention}(Q,K,V)=\mathrm{Softmax}(\frac{Q K^T}{\sqrt{d}}) \cdot V$, with 
{\small 
$$Q=W^Q z_{v_i}, K=W^K z_{v_i}, V=W^V z_{v_i},$$}%
where $W^Q$, $W^K$, and $W^V$ are learnable matrices that project the inputs to query, key and value, respectively, and and $d$ is the output dimension of key and query features. 

We extend a 2D LDM to the spatio-temporal domain. Similar to VDM~\cite{ho2022video}, we inflate the 2D convolution layers to pseudo 3D convolution layers, with $3\times3$ kernels being replaced by $1\times3\times3$ kernels and append a temporal self-attention layer in each transformer block for temporal modeling. To enhance the temporal coherence, we further extend the spatial self-attention mechanism to the spatio-temporal domain. There are alternative options for spatio-temporal attention (ST-Attn) mechanism, including full attention and causal attention which also capture spatio-temporal consistency. However, such straightforward choices are actually not feasible in generating videos with increasing frames due to their high computational complexity. Specifically, given $m$ frames and $N$ sequences for each frame, the complexity for both full attention and causal attention is $\mathcal{O}((mN)^2)$. It is not affordable if we need to generate long videos with a large value of $m$.

Here, we propose to use a sparse version of causal attention mechanism, where the attention matrix are computed between frame $z_{v_i}$ and two previous frames $z_{v_1}$ and $z_{v_{i-1}}$, remaining low computational complexity at $\mathcal{O}(2m(N)^2)$. Specifically, we derive query feature from frame $z_{v_i}$, key and value features from the \textit{first} frame $z_{v_1}$ and the \textit{former} frame $z_{v_{i-1}}$, and implement $\mathrm{Attention}(Q,K,V)$ with
{\small
$$Q=W^Q z_{v_i}, K=W^K \left[ z_{v_1}, z_{v_{i-1}} \right], V=W^V \left[ z_{v_1}, z_{v_{i-1}} \right],$$}%
where $\left[ \cdot \right]$ denotes concatenation operation. Note that the projection matrices $W^Q$, $W^K$, and $W^V$ are shared across space and time. See Fig.~\ref{fig:our-attn} for a visual depiction.

\subsection{Fine-Tuning and Inference}\label{sec:finetune}

\paragraph{Model fine-tuning.}
We now finetune our network on the given input video for temporal modeling. The spatio-temporal attention (ST-Attn) is designed to model temporal consistency by querying relevant positions in previous frames. Therefore, we propose to fix parameters $W^K$ and $W^V$, and only update $W^Q$ in ST-Attn layers. In contrast, we finetune the entire temporal self-attention (T-Attn) layers as they are newly added. 
Moreover, we propose to refine the text-video alignment by updating the query projection in cross-attention (Cross-Attn). In practice, finetuning the attention blocks is computationally efficient compared to full tuning~\cite{ruiz2022dreambooth}, and meanwhile retains the original property of pre-trained T2I diffusion models. We use the same training objective in standard LDMs~\cite{rombach2022high}. Fig.~\ref{fig:arch} illustrates the finetuning process with the trainable parameters highlighted.

\vspace{-1em}
\paragraph{Structure guidance via DDIM inversion.} 
Finetuning the attention layers is essential to ensure spatial consistency across all frames. However, it does not offer much control over pixel shifts, resulting in stagnant videos in the loop. To tackle this problem, we incorporate structure guidance from the source video during the inference stage. Specifically, we obtain a latent noise of source video $\mathcal{V}$ through DDIM inversion with no textual condition. This noise serves as the starting point for DDIM sampling, which is guided by an edited prompt $\mathcal{T^\ast}$. The output video $\mathcal{V^\ast}$ is then given by 
% \\[-10pt]
\begin{equation*}
\mathcal{V^\ast} = \mathcal{D}(\text{DDIM-samp}(\text{DDIM-inv}(\mathcal{E}(\mathcal{V})), \mathcal{T^\ast})).
\end{equation*}
% \\[-10pt]
Note that for the same input video, we only need to perform DDIM inversion once. Our experiments demonstrate its effectiveness in accurately conveying the structural movements from the source video to the generated videos.

\begin{figure*}[t!]
    \centering
    \includegraphics[width=0.99\linewidth]{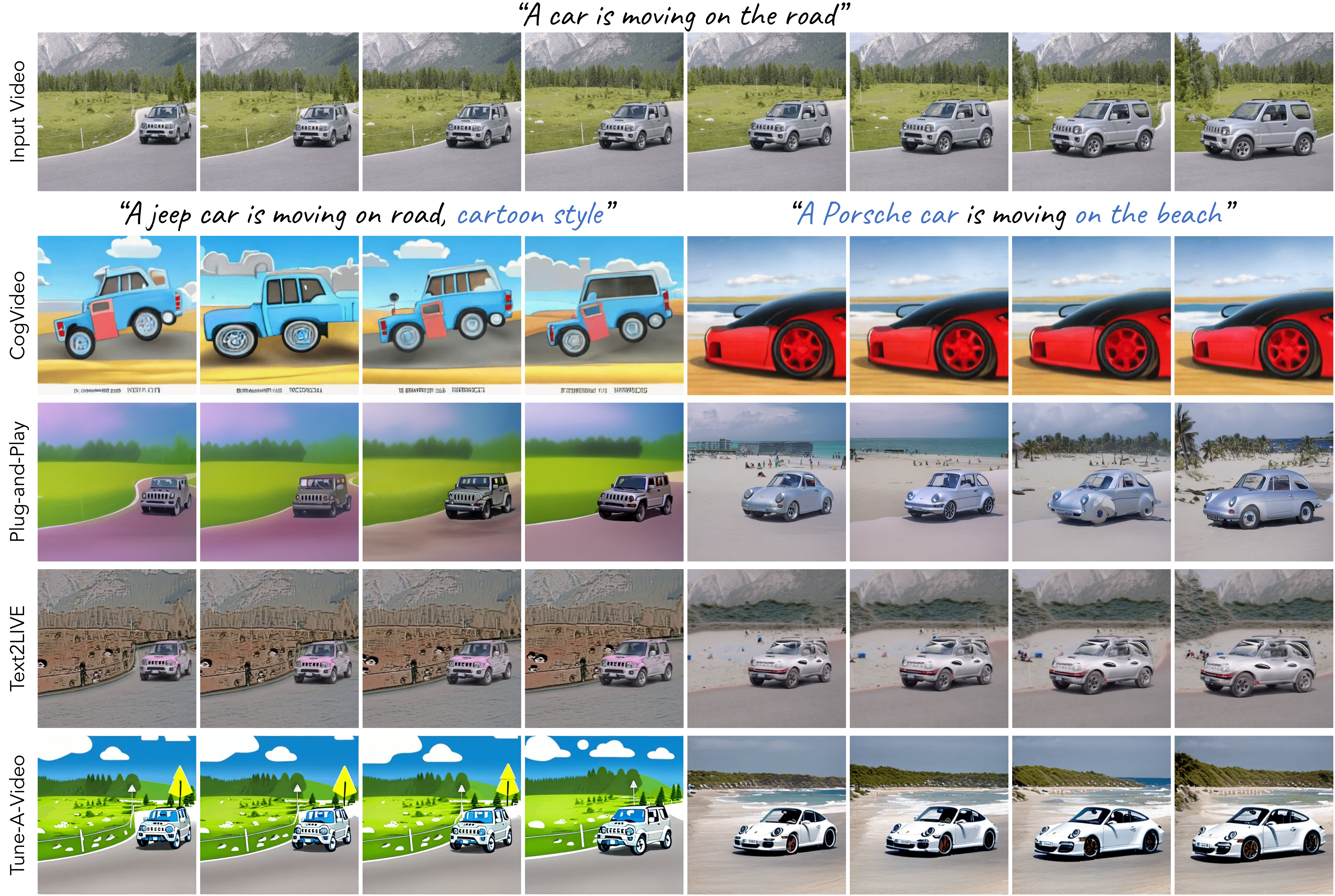}
    \caption{\textit{\textbf{Qualitative comparison between evaluated methods.}} Zoom in for best view.}
    \label{fig:baselines}
    \vspace{-1em}
\end{figure*}

\section{Applications of \model{}}\label{sec:app}

We showcase several applications of our \model{} for text-driven video generation and editing. %(see Fig.~\ref{fig:teaser}, Fig.~\ref{fig:results}, Fig.~\ref{fig:suppl-results-1} and Fig.~\ref{fig:suppl-results-2}). 

\vspace{-1em}
\paragraph{Object editing.}
One of the major applications of our method is to modify the object through the editing of text prompts. This allows replacing, adding, or removing objects with ease. Fig.~\ref{fig:results} shows some examples. 
We can replace ``a man" with ``Spider Man" or ``Wonder Woman", ``a rabbit" with ``a cat" or ``a puppy", or even switch out ``a watermelon" for ``a cheeseburger", simply by modifying the corresponding words. We can add an object such as ``a cowboy hat" or ``sunglasses" by further describing it in the prompt. To remove an object, we can easily delete the corresponding phrase---for example, the watermelon.

\vspace{-1em}
\paragraph{Background change.} 
Our method also enables users to change the video background (\ie, the place where the object is), while preserving the consistency of the object's movements. For example, we can modify the background of the skiing man in Fig.~\ref{fig:results} to be ``on the beach" or ``at sunset", by adding a new location/time description, and change the countryside road view in Fig.~\ref{fig:baselines} to sea view, by replacing an existing location description. 

\vspace{-1em}
\paragraph{Style transfer.} 
Thanks to the open-domain knowledge of pretrained T2I models, our method transfer videos into a variety of styles that are difficult to learn solely from video data \cite{singer2022make}. For example, we transform real-world videos into comic styles (Fig.~\ref{fig:results}), or Van Gogh style (Fig.~\ref{fig:suppl-results-1}), by appending the global style descriptor to the prompt. 

\vspace{-1em}
\paragraph{Personalized and controllable generation.} 

Our method can be easily integrated with personalized T2I models (\eg, DreamBooth~\cite{ruiz2022dreambooth}, which takes 3-5 images as input and returns a personalized T2I model), by directly finetuing on them.  For instance, we can use a DreamBooth personalized for ``Modern Disney Style" or ``Mr Potato Head" to create videos of a specific style or subject (Fig.~\ref{fig:suppl-results-2}). Our method can also be integrated with conditional T2I models like T2I-Adapter~\cite{mou2023t2i} and ControlNet~\cite{zhang2023adding}, to enable diverse controls on the generated videos at no extra training cost. For example, we can further edit the motion using a sequence of human pose as control (\eg, dancing in Fig.~\ref{fig:teaser}). Note that the human pose sequence can be automatically detected from real-world videos using an off-the-shelf pose estimation model~\cite{cao2017realtime}. The compatibility of our method with personalized and conditional T2I models offers more possibilities for users to create the video content they desire.

\section{Experiments}

\subsection{Implementation Details}

Our development is based on Latent Diffusion Models~\cite{rombach2022high} (a.k.a Stable Diffusion) and the public pretrained weights\footnote{https://huggingface.co/CompVis/stable-diffusion-v1-4}. 
We sample $32$ uniform frames at resolution of $512 \times 512$ from input video, and finetune the models with our method for $500$ steps on a learning rate $3 \times 10^{-5}$ and a batch size $1$.
At inference, we use DDIM sampler~\cite{song2020denoising} with classifier-free guidance~\cite{ho2022classifier} in our experiments. For a single video, it takes about $10$ minutes for finetuning, and about $1$ minute for sampling on a NVIDIA A100 GPU.

\subsection{Baseline Comparisons}

\paragraph{Dataset.} 

To evaluate our approach, we use 42 representative videos taken from DAVIS dataset~\cite{pont20172017}. We automatically produce the video footage using an off-the-shelf captioning model~\cite{li2023blip}, and manually design 140 edited prompts across our applications in Sec.~\ref{sec:app}. More details on our benchmark are provided in Sec.~\ref{sec:dataset}.

\vspace{-1em}
\paragraph{Baselines.} 

We compare our method against three baselines: 1) \textit{CogVideo}~\cite{hong2022cogvideo}: a T2V model trained on a dataset of 5.4 million captioned videos, and is capable of generating videos directly from text prompts in a zero-shot manner. 2) \textit{Plug-and-Play}~\cite{tumanyan2022plug}: a cutting-edge image editing model that can edit each frame of a video individually. 3) \textit{Text2LIVE}~\cite{bar2022text2live}: a recent approach for text-guided video editing that employs layered neural atlases~\cite{kasten2021layered}. 

\vspace{-1em}
\paragraph{Qualitative results.}

We present a visual comparison of our approach against several baselines in Fig.~\ref{fig:baselines}. We observe that while CogVideo can produce videos that reflect the general concept in the text, the output videos varies a lot in quality and it cannot take a video as input. Plug-and-Play, on the other hand, successfully edits each video frame individually, but lacks frame consistency as the temporal context is neglected (\eg, the appearance of the Porsche car is not consistent across frames). Text2LIVE, while capable of producing temporally smooth videos, struggles to accurately represent the edited prompt (\eg, the Porsche car still appears in the shape of the original jeep car). This may be due to its reliance on layered neural atlases, which restricts its editing ability. In contrast, our method generates temporally-coherent videos that preserve structural information from the input video and align well with edited words and details. Additional qualitative comparison can be found in Fig.~\ref{fig:suppl-baselines}.

\vspace{-1em}
\paragraph{Quantitative results.} 

\begin{table}[]
\centering
\caption{\textit{\textbf{Quantitative comparison with evaluated baselines.}} * indicates \model{} \vs CogVideo, ** indicates \model{} \vs Plug-and-Play.}
\resizebox{\columnwidth}{!}{%
\begin{tabular}{lccccc}
\hline
\multirow{2}{*}{Method} & \multicolumn{2}{c}{Frame Consisitency}  &  & \multicolumn{2}{c}{Textual alignment}   \\ \cline{2-3} \cline{5-6} 
              & CLIP Score & User Preference &  & CLIP Score & User Preference \\ \hline
CogVideo      & 90.64      & 12.14           &  & 23.91      & 15.00           \\
Plug-and-Play & 88.89      & 37.86           &  & 27.56      & 23.57           \\
Tune-A-Video            & \textbf{92.40} & \textbf{87.86* / 62.14**} &  & \textbf{27.58} & \textbf{85.00* / 76.43**} \\ \hline
\end{tabular}%
}
\label{tab:quant}
% \vspace{-1em}
\end{table}

\begin{figure}[t!]
    \centering
    \includegraphics[width=0.99\linewidth]{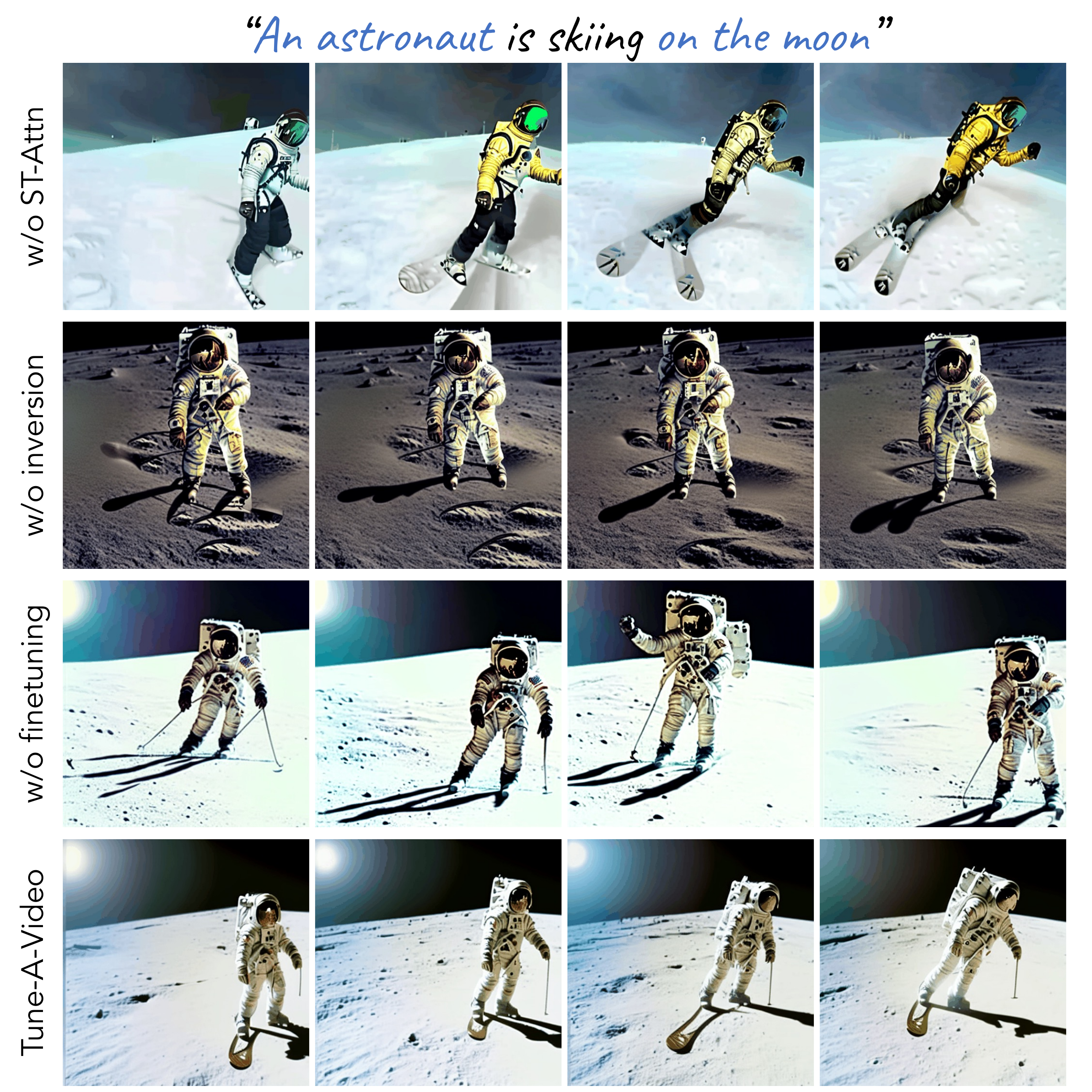}
    \caption{\textit{\textbf{Ablation study.}} Input video is shown in Fig~.\ref{fig:results}.}
    \label{fig:ablation}
    \vspace{-1em}
\end{figure}

We quantify our method against baselines through automatic metrics and user study, and report frame consistency and textual faithfulness in Tab.~\ref{tab:quant}.

\textit{Automatic metrics.} For frame consistency, we compute CLIP~\cite{radford2021learning} image embeddings on all frames of output videos and report the average cosine similarity between all pairs of video frames. To measure textual faithfulness, we compute average CLIP score between all frames of output videos and corresponding edited prompts. 
Our results indicate that CogVideo produces consistent video frames but struggle to represent the textual description, whereas Plug-and-Play achieves high textual faithfulness but failed to generate consistent content. In contrast, our method outperforms baselines in both metrics. 

\textit{User study.} For frame consistency, we present two videos generated by our method and a baseline in random order and ask the raters ``which one has better temporal consistency?". For textual faithfulness, we additionally show the textual description and ask the raters ``which video better aligns with the textual description?". We recruit 5 participants to annotate each example and use a majority vote for the final result. Additional details are provided in Appendix (Sec.~\ref{sec:user-study}). We observe that CogVideo and Plug-and-Play are less preferred due to frame-wise and frame-text inconsistency, whereas our method achieves higher user preference in both aspects. 

\subsection{Ablation Study}

We conduct an ablation study to assess the importance of the spatio-temporal attention (ST-Attn) mechanism, DDIM inversion, and finetuning in our \model{}. Each design is individually ablated to analyze its impact. The results, presented in Fig.~\ref{fig:ablation}, show that the model \textit{w/o ST-Attn} displays significant content discrepancies (evident from the skier's clothing color). In contrast, the model \textit{w/o inversion} maintains consistent content but fails to replicate the motion (\ie, skiing) in the input video. Thanks to the ST-Attn and inversion, model \textit{w/o finetuning} still suffices consistent content across frames. However, the motion in consecutive frames is not smooth, resulting in flickering videos. Additional video examples of ablation study can be found in Fig.~\ref{fig:suppl-ablation}. These results indicate that all of our key designs contribute to the successful results of our method.

\section{Limitations and Future Work}

\begin{figure}[t!]
    \centering
    \includegraphics[width=0.99\linewidth]{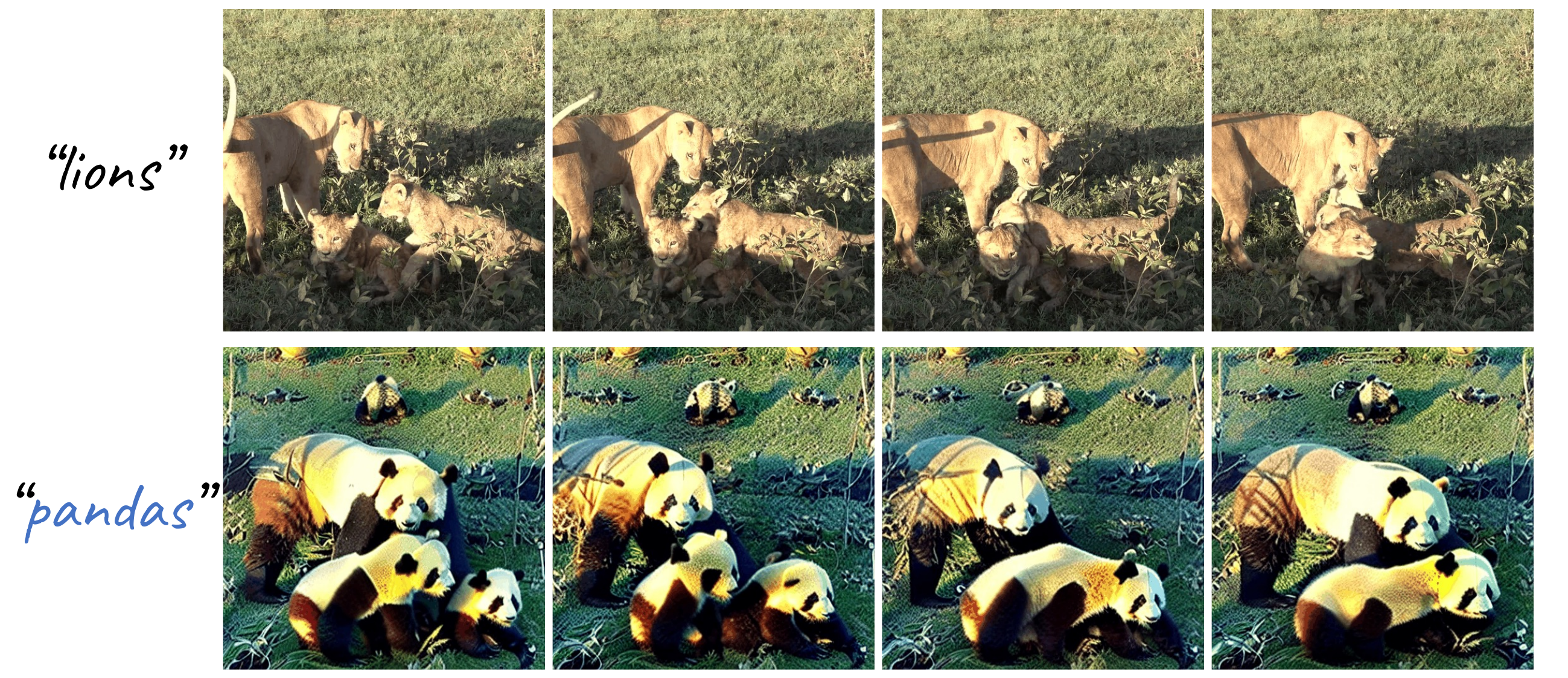}
    \caption{\textit{\textbf{Limitations:}} Our method might produce unpleasant results when the input video contains multiple objects and exhibits occlusions. For example, the two pandas at the bottom being mixed together.}
    \label{fig:failure}
    \vspace{-1em}
\end{figure}

Fig.~\ref{fig:failure} presents a failure case of our method when the input video contains multiple objects and exhibits occlusion. This may be due to the inherent limitation of the T2I model in handling multiple objects and object interactions. A potential solution is to use additional conditional information, such as depth, to enable the model to differentiate between different objects and their interactions. This avenue of research is left as future work.

\section{Conclusion}

In this paper, we introduce a new task for T2V generation called \setting{}. This task involves training a T2V generator using only a single text-video pair and pretrained T2I models. We present \model{}, a simple yet effective framework for text-driven video generation and editing. To generate continuous videos, we propose an efficient tuning strategy and structural inversion that enable generating temporally-coherent videos. Extensive experiments demonstrate the remarkable results of our method spanning a wide range of applications.

% \clearpage
% \newpage
{\small
% \bibliographystyle{ieee_fullname}
% \bibliography{refs}

}

\onecolumn
% \appendix
% \vspace{-2cm}
% \begin{center}
% \LARGE\textbf{Appendix}
% \end{center}

\clearpage
\appendix
% \onecolumn
\appendixpage

\section{Dataset Details}\label{sec:dataset}

We select 42 videos from the DAVIS dataset~\cite{pont20172017}, covering a range of categories including animals, vehicles, and humans. The selected video items are listed in Tab.~\ref{tab:dataset}. To obtain video footage, we use BLIP-2~\cite{li2023blip} for automated captions. We then manually design three edited prompts for each video, resulting 140 edited prompts in total. These edited prompts include object editing, background changes, and style transfers, as described in Sec.~\ref{sec:app}. 

\begin{table}[h]
\centering
\caption{\textit{\textbf{Names of videos selected from DAVIS dataset.}}}
\label{tab:dataset}
\begin{tabular}{|c|c|c|}
\hline
bear & blackswan & boat \\
breakdance-flare & camel & car-roundabout \\
car-shadow & car-turn & cows \\
dog & dog-agility & drift-turn \\
elephant & flamingo & girl-dog \\
gold-fish & golf & guitar-violin \\
hike & hockey & horsejump-high \\
horsejump-low & kid-football & kite-surf \\
lab-coat & libby & lions \\
longboard & lucia & mallard-water \\
man-bike & mbike-santa & mbike-trick \\
motorbike & parkour & rhino \\
running & scooter-gray & snowboard \\
swing & tandem & tennis \\
\hline
\end{tabular}
\end{table}

\section{User Study Details}\label{sec:user-study}

We conduct a user study on our dataset of 140 edited prompts to compare our method against two baselines: Plug-and-Play~\cite{tumanyan2022plug} and CogVideo~\cite{hong2022cogvideo}. The comparison results are shown in Tab.~\ref{tab:quant}. The participants of the user study are mainly students and colleagues in university. We ask 5 raters to evaluate each edited prompt by comparing two videos generated by two different methods (shown in random order) and answering two following questions:

\begin{enumerate}
    \item Which video has higher consistency? Please select the one that looks more smooth as a video.
    \item Which video matches the text better? Please select the one that better represents the given text description.
\end{enumerate}

\section{Additional Results}

\begin{figure*}[t!]
    \centering
    \includegraphics[width=0.99\linewidth]{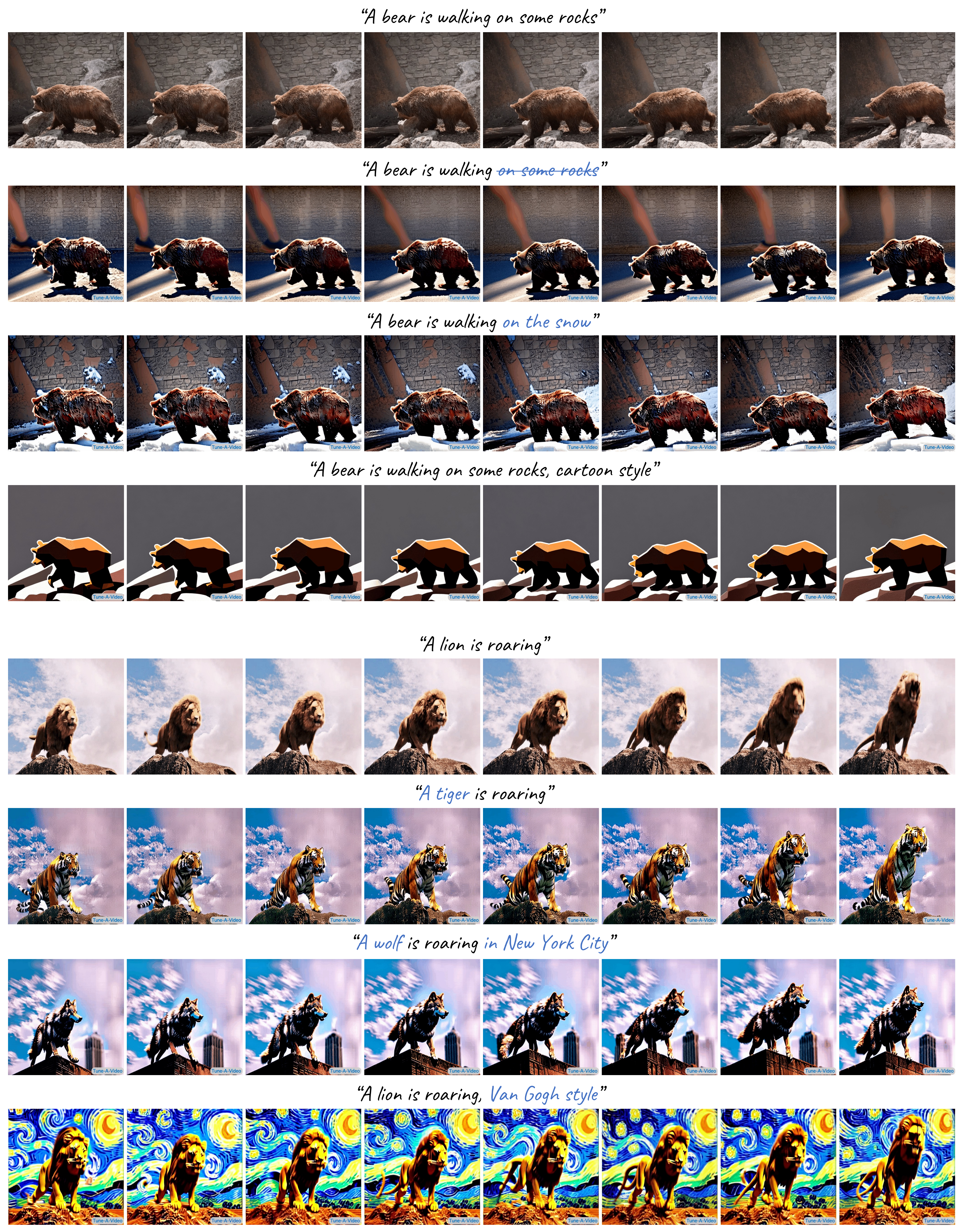}
    \caption{\textit{\textbf{Additional sample results of our method (1/2).}}}
    \label{fig:suppl-results-1}
\end{figure*}

\begin{figure*}[t!]
    \centering
    \includegraphics[width=0.99\linewidth]{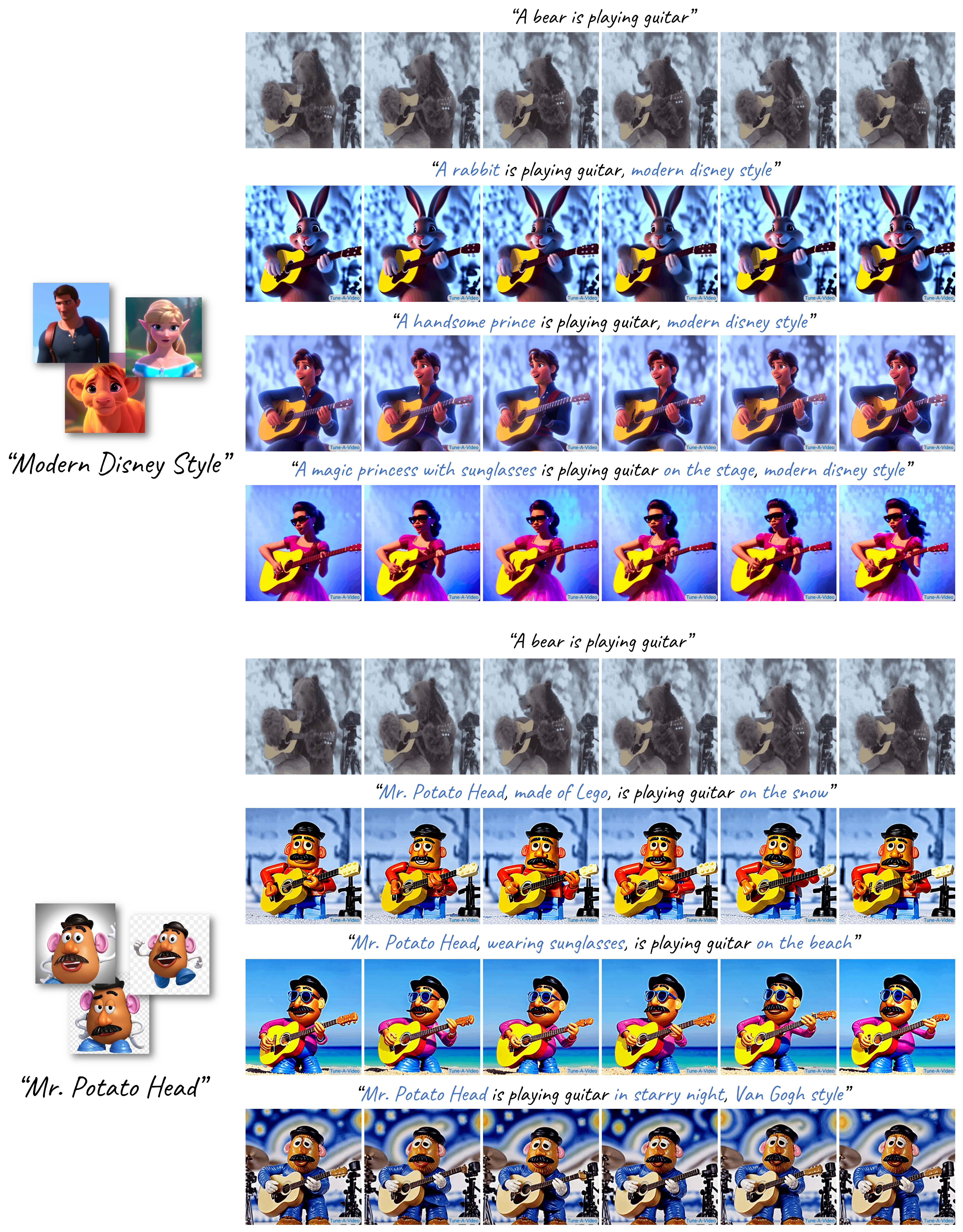}
    \caption{\textit{\textbf{Additional sample results of our method (2/2).}}}
    \label{fig:suppl-results-2}
\end{figure*}

\begin{figure*}[t!]
    \centering
    \includegraphics[width=0.99\linewidth]{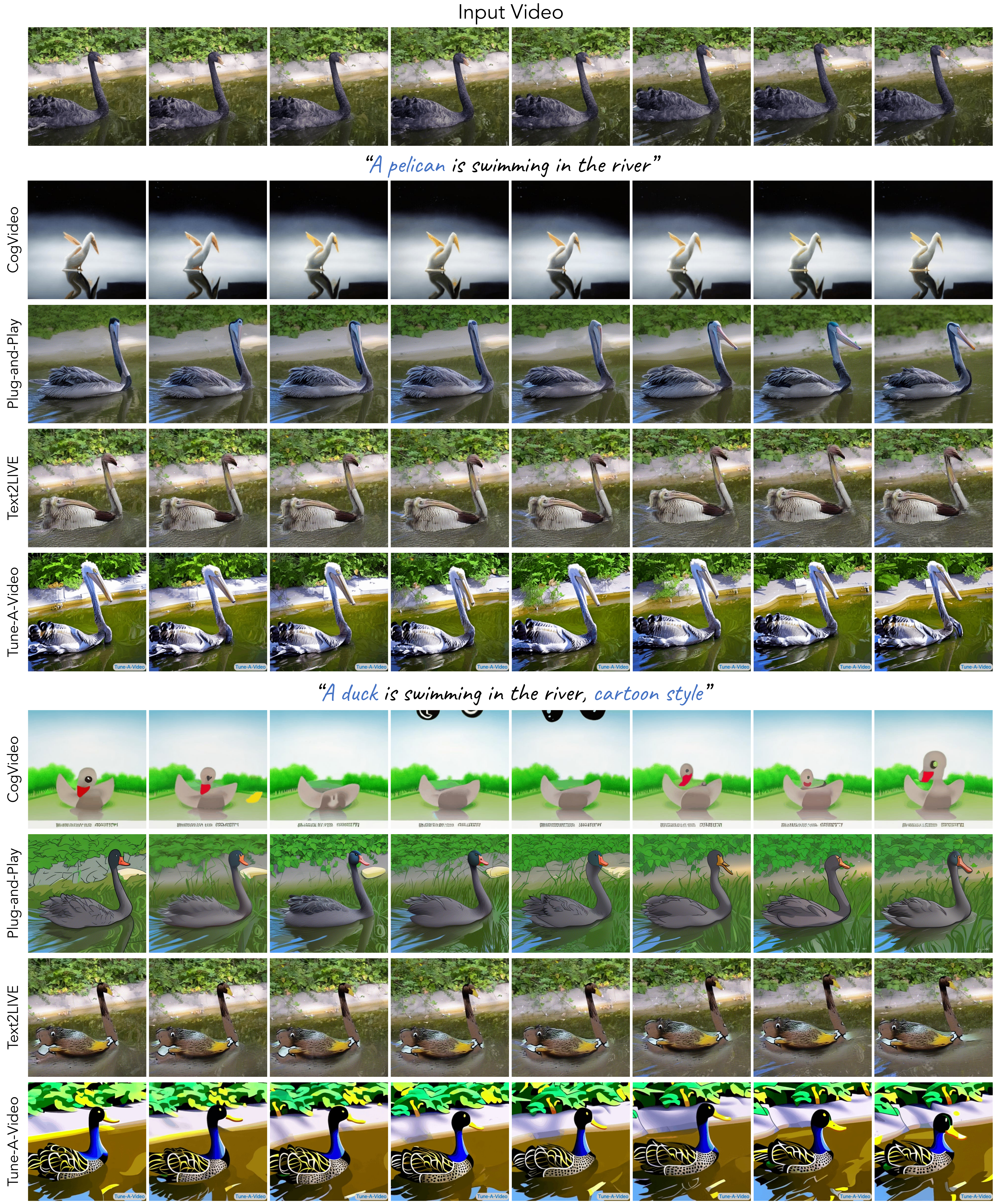}
    \caption{\textit{\textbf{Additional qualitative comparsion between evaluated methods.}}}
    \label{fig:suppl-baselines}
\end{figure*}

\begin{figure*}[t!]
    \centering
    \includegraphics[width=0.99\linewidth]{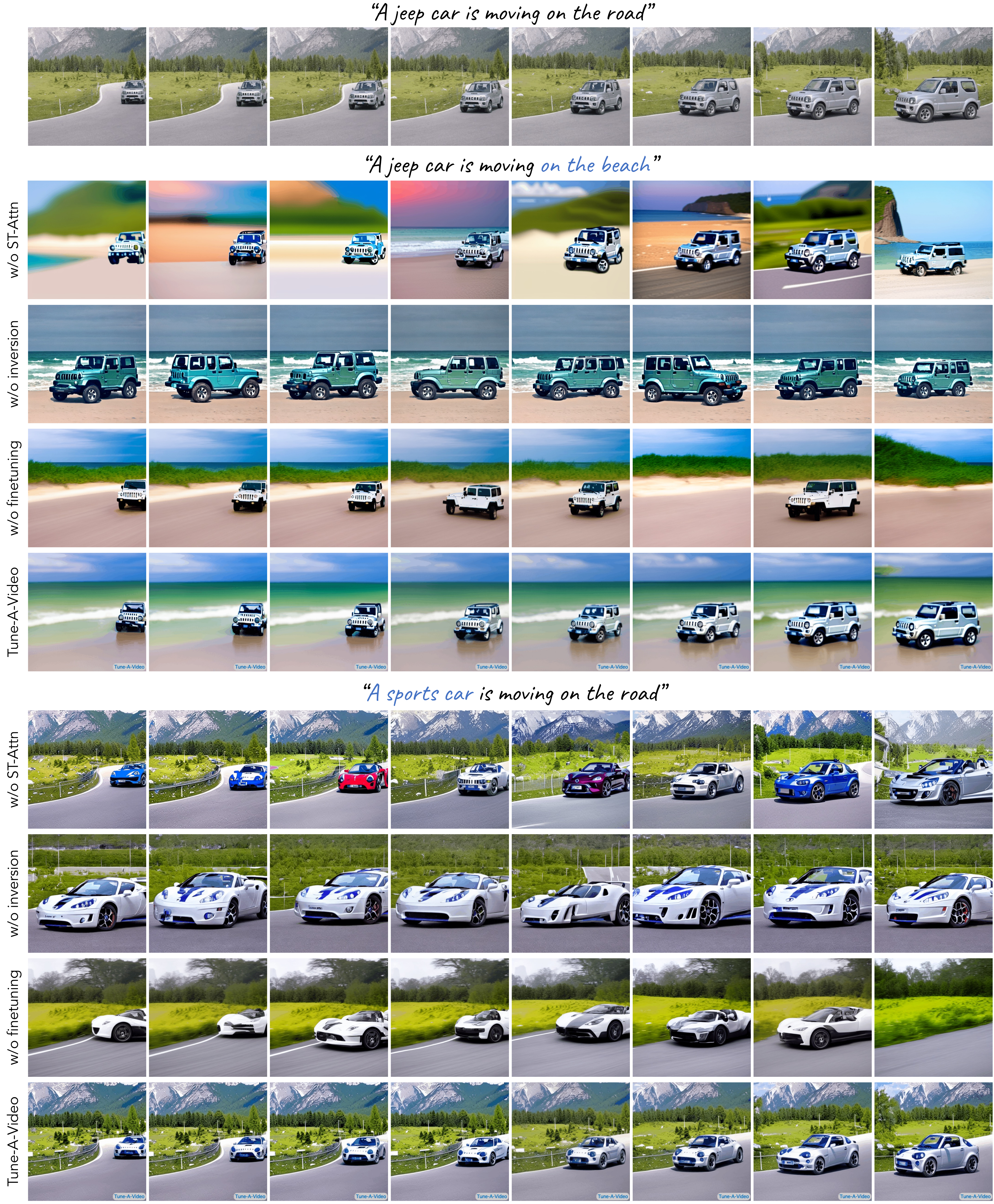}
    \caption{\textit{\textbf{Additional ablation study.}}}
    \label{fig:suppl-ablation}
\end{figure*}

Fig.~\ref{fig:suppl-results-1} and Fig.~\ref{fig:suppl-results-2} showcase additional video examples of our methods, Fig.~\ref{fig:suppl-baselines} provides additional comparison with baselines, and Fig.~\ref{fig:suppl-ablation} gives additional results of ablation study. 

\end{document}